\documentclass[11pt]{article}

\usepackage[utf8]{inputenc}
\usepackage[T1]{fontenc}
\usepackage{times}
\usepackage[top=0.85in, bottom=0.85in, left=0.75in, right=0.75in]{geometry}
\usepackage{graphicx}
\usepackage{booktabs}
\usepackage{amsmath}
\usepackage{hyperref}
\usepackage{xcolor}
\usepackage{caption}
\usepackage{natbib}
\usepackage{enumitem}
\usepackage{microtype}
\usepackage{titlesec}
\usepackage{multicol}
\usepackage{colortbl}

\definecolor{darkblue}{RGB}{0,51,102}
\definecolor{medblue}{RGB}{30,80,140}
\definecolor{lightblue}{RGB}{220,235,250}
\definecolor{lightgray}{RGB}{245,245,245}
\definecolor{tableheader}{RGB}{210,225,245}

\hypersetup{colorlinks=true, linkcolor=darkblue, citecolor=medblue, urlcolor=medblue}

\titleformat{\section}{\normalfont\large\bfseries\color{darkblue}}{\thesection.}{0.5em}{}[\vspace{0.1em}\titlerule]
\titleformat{\subsection}{\normalfont\normalsize\bfseries\color{medblue}}{\thesubsection}{0.5em}{}
\titlespacing*{\section}{0pt}{14pt plus 2pt}{6pt plus 1pt}
\titlespacing*{\subsection}{0pt}{10pt plus 2pt}{4pt plus 1pt}

\title{\vspace{-0.3cm}\rule{\textwidth}{2.5pt}\\[0.35cm]{\fontsize{16}{19}\selectfont\textbf{Implicit Grading Bias in Large Language Models:\\[0.15cm]How Writing Style Affects Automated Assessment\\[0.15cm]Across Math, Programming, and Essay Tasks}}\\[0.25cm]\rule{\textwidth}{0.6pt}\vspace{0.1cm}}

\author{\textbf{Rudra Jadhav}\\[0.1cm]{\small Department of Computer Science}\\{\small Savitribai Phule Pune University}\\{\small Pune, India}\\{\small\texttt{roodra.jadhav@gmail.com}}\and\textbf{Janhavi Danve}\\[0.1cm]{\small Department of Computer Science}\\{\small Savitribai Phule Pune University}\\{\small Pune, India}\\{\small\texttt{janhavi.danve@gmail.com}}\and\textbf{Sonalika Shaw}\\[0.1cm]{\small Department of Computer Science}\\{\small Dr.\ D.\ Y.\ Patil School of}\\{\small Science and Technology}\\{\small Pune, India}\\{\small\texttt{shawsonalika@gmail.com}}}

\date{\small\textcolor{gray}{March 2026 \quad$\vert$\quad Preprint --- Under Review}}

\begin{document}
\maketitle
\thispagestyle{empty}

\vspace{-0.2cm}
\noindent\colorbox{lightgray}{\begin{minipage}{\dimexpr\textwidth-2\fboxsep}
\vspace{0.3cm}
\centerline{\textbf{\textcolor{darkblue}{Abstract}}}
\vspace{0.15cm}
\small
As large language models (LLMs) are increasingly deployed as automated graders in educational settings, concerns about fairness and bias in their evaluations have become critical. This study investigates whether LLMs exhibit implicit grading bias based on writing style when the underlying content correctness remains constant. We constructed a controlled dataset of 180 student responses across three subjects (Mathematics, Programming, and Essay/Writing), each with three surface-level perturbation types: grammar errors, informal language, and non-native phrasing. Two state-of-the-art open-source LLMs --- LLaMA 3.3 70B (Meta) and Qwen 2.5 72B (Alibaba) --- were prompted to grade responses on a 1--10 scale with explicit instructions to evaluate content correctness only and to disregard writing style.

\vspace{0.15cm}
Our results reveal statistically significant grading bias in Essay/Writing tasks across both models and all perturbation types ($p < 0.05$), with effect sizes ranging from medium (Cohen's $d = 0.64$) to very large ($d = 4.25$). Informal language received the heaviest penalty, with LLaMA deducting an average of \textbf{1.90 points} and Qwen deducting \textbf{1.20 points} on a 10-point scale --- penalties comparable to the difference between a B+ and C+ letter grade. Non-native phrasing was penalized 1.35 and 0.90 points respectively. In sharp contrast, Mathematics and Programming tasks showed minimal bias, with most conditions failing to reach statistical significance. These findings demonstrate that LLM grading bias is subject-dependent, style-sensitive, and persists despite explicit counter-bias instructions in the grading prompt. We discuss implications for equitable deployment of LLM-based grading systems and recommend bias auditing protocols before institutional adoption.

\vspace{0.2cm}
\noindent\textbf{\textcolor{darkblue}{Keywords:}} \textit{LLM bias, automated grading, educational assessment, writing style, non-native phrasing, fairness, implicit bias, open-source language models, educational equity}
\vspace{0.2cm}
\end{minipage}}

\vspace{0.3cm}

\begin{multicols}{2}

\section{Introduction}

The integration of large language models into educational assessment represents one of the most consequential applications of artificial intelligence in the public sector~\cite{guide2026, gradeopt2024, cohn2024}. Universities and schools worldwide face chronic grading bottlenecks: large class sizes, limited teaching assistant availability, and the inherent subjectivity of evaluating open-ended responses. LLMs offer an attractive solution --- they can process thousands of student responses in minutes, provide individualized feedback, and operate continuously without fatigue. Platforms such as Coursera, Khan Academy, and Duolingo have already begun incorporating LLM-based assessment tools into their workflows, signaling a broader institutional shift toward automated evaluation.

However, the promise of automated grading rests on a critical and often untested assumption: that these models evaluate student work fairly, regardless of the student's linguistic background, writing conventions, or cultural context~\cite{haim2024, kwako2024, weissburg2025}. In real-world classrooms, students do not write with uniform style. Students with non-native phrasing patterns may produce grammatically unconventional but conceptually correct responses. Students from informal educational backgrounds may express valid ideas using colloquial language. First-generation university students may lack exposure to the formal academic register that is over-represented in LLM training data.

If LLMs penalize these surface-level variations while claiming to evaluate content correctness, they effectively discriminate against the very students who stand to benefit most from scalable, affordable assessment tools. Rather than democratizing education, biased automated grading would reinforce existing inequalities --- giving higher scores to students who already write in the polished style the model was trained on, and penalizing those who express the same knowledge differently.

This paper investigates whether such implicit bias exists, how severe it is, and which academic subjects are most vulnerable. Our study makes three primary contributions:

\begin{enumerate}[leftmargin=*, topsep=3pt, itemsep=2pt]
\item We design a \textbf{controlled perturbation framework} that isolates writing style from content correctness, enabling direct measurement of surface-level bias.
\item We evaluate two state-of-the-art open-source LLMs across \textbf{three distinct academic domains}, revealing a sharp contrast between objective and subjective grading.
\item We demonstrate that \textbf{explicit prompt-level instructions} to ignore writing style are insufficient to prevent bias, raising fundamental questions about prompt engineering as a debiasing strategy for high-stakes educational applications.
\end{enumerate}

\section{Related Work}

\subsection{LLMs as Automated Graders}

A growing body of work has explored whether LLMs can serve as reliable automated graders. The GUIDE framework~\cite{guide2026} addressed the challenge of selecting effective few-shot examples for grading prompts, showing that a small number of carefully chosen exemplars --- particularly those near scoring boundaries --- can substantially improve rubric adherence across multiple STEM disciplines. The key insight is that \textit{which} examples are shown to the model matters as much as \textit{how many}.

Building on this line of work, the GradeOpt system~\cite{gradeopt2024} introduced a self-improving pipeline where the grading model identifies its own errors and iteratively refines its scoring guidelines. Their results suggest that automated guideline refinement can produce scoring criteria that align with human judgment more closely than manually authored rubrics --- an encouraging result for scalability.

Broader evaluations have compared LLM grading against human baselines in university settings. One study across bioinformatics coursework~\cite{bioinfo2025} reported that freely available open-source models matched the scoring accuracy of commercial APIs under controlled prompting conditions. Another comparative study~\cite{whogrades2025} observed that LLM-generated grades tend to skew higher than instructor grades, and that models are better at recognizing strong work than at distinguishing among weaker submissions --- a pattern that informed our decision to use varied ground-truth scores (7--10) rather than uniformly high baselines.

\subsection{Bias in LLM-Based Education}

The question of whether LLMs treat all students equally has received growing attention. The most extensive investigation to date~\cite{weissburg2025} examined how nine different LLMs tailor educational explanations for students from varying demographic backgrounds. Testing across multiple sensitive attributes --- including socioeconomic status and disability --- the authors reported measurable disparities in how models adjusted content difficulty and framing for different groups. A key limitation of that study, which our work directly addresses, is its focus on how LLMs \textit{teach} rather than how they \textit{evaluate}. The grading context introduces distinct bias risks because scores carry direct consequences for students.

Separately, research on automated essay scoring~\cite{kwako2024} has shown that language models pick up on demographic cues embedded in writing style --- even when no demographic labels are provided. This suggests that the way a student writes can serve as a proxy for their background, potentially triggering differential scoring. Our work operationalizes this insight directly: rather than providing demographic labels, we let writing style itself carry the signal and measure whether it distorts scores.

\subsection{Broader LLM Bias Research}

Beyond education, a substantial literature documents systematic bias in LLM-generated decisions. Audit studies have revealed that when models are given scenarios involving named individuals, the quality of advice and recommendations shifts depending on whether the name signals a majority or minority demographic~\cite{haim2024}. This effect has been replicated in hiring contexts, where resume-ranking experiments across hundreds of job listings and millions of pairwise comparisons showed stark disparities along racial and gender lines~\cite{wilson2024}. In the educational domain specifically, controlled comparisons of LLM-based student selection decisions~\cite{audit2025} have shown that bias magnitude and even direction can differ substantially between model versions, underscoring the need for version-specific auditing rather than blanket trust in any single model family.

Our study occupies the intersection of these two research trajectories. Where prior grading research has focused on accuracy and prior bias research has focused on explicit demographic cues (names, labels), we test whether the natural linguistic variation in student writing --- the way people actually differ in educational settings --- is sufficient to trigger biased evaluation by LLMs.

\section{Methodology}

\subsection{Dataset Construction}

We constructed a controlled dataset comprising 60 unique questions distributed equally across three subjects: Mathematics (20 questions covering algebra, geometry, statistics, and arithmetic), Programming (20 Python coding tasks spanning basic functions, algorithms, and data manipulation), and Essay/Writing (20 argumentative essay prompts on contemporary social topics). For each question, we authored a base answer representing a correct, well-structured response in standard academic English.

Each base answer was then systematically perturbed along three dimensions while strictly preserving content correctness:

\begin{itemize}[leftmargin=*, topsep=3pt, itemsep=3pt]
\item \textbf{Grammar Errors:} Introduction of spelling mistakes, punctuation errors, and grammatical inconsistencies typical of unpolished writing (e.g., missing articles, incorrect subject-verb agreement, run-on sentences).
\item \textbf{Informal Language:} Conversion of formal prose into casual, conversational style using contractions, slang, and colloquial expressions while maintaining identical logical content (e.g., ``\textit{u gotta subtract 5 from both sides so 2x = 8 then just divide by 2 and boom x = 4}'').
\item \textbf{Non-native Phrasing:} Reformulation using patterns characteristic of non-native English speakers, including article misuse, atypical preposition selection, and direct translation artifacts from common L1 languages (e.g., ``\textit{We subtract 5 from both the sides: 2x is becoming 8}'').
\end{itemize}

This $3 \times 3 \times 20$ factorial design yielded 180 perturbed responses, each paired with its original base answer for within-subject comparison. Human ground-truth scores were assigned to each question on a 1--10 scale based on content correctness, completeness, and reasoning depth, ranging from 7 to 10 across the dataset (Math mean: 9.0; Programming mean: 8.75; Essay mean: 8.45). The deliberate variation in ground-truth scores enables analysis of whether bias intensity correlates with answer quality level.

\subsection{Models Under Evaluation}

We evaluated two open-source large language models of comparable scale but distinct training lineages:

\vspace{0.15cm}
\noindent\textbf{LLaMA 3.3 70B Instruct} (Meta): A 70-billion parameter model from Meta's LLaMA family, accessed via the Groq inference platform. LLaMA 3.3 represents Meta's state-of-the-art open-source offering, trained predominantly on English-dominant web corpora with extensive instruction tuning for conversational and analytical tasks.

\vspace{0.1cm}
\noindent\textbf{Qwen 2.5 72B Instruct} (Alibaba Cloud): A 72-billion parameter model from Alibaba's Qwen family, accessed via the HuggingFace Inference API. Qwen 2.5 was trained on a significantly more multilingual corpus with substantial representation of Chinese, Southeast Asian, and other non-English languages, providing a meaningful contrast in training data composition and cultural embedding.

\vspace{0.15cm}
The selection of these specific models was deliberate: both are large-scale, instruction-tuned, open-source models of nearly identical parameter count, but they differ substantially in their training data distribution, organizational origin (American vs.\ Chinese), and linguistic diversity. This design enables us to assess whether observed bias patterns reflect model-specific artifacts or systemic properties of large-scale language modeling.

\subsection{Grading Protocol}

Each model received an identical standardized grading prompt for every response, following established best practices in LLM-based assessment~\cite{guide2026, gradeopt2024, cohn2024}. The prompt specified the academic subject, presented the question and student answer verbatim, and provided a detailed rubric mapping scores 1--10 to qualitative performance levels (10 = perfectly correct and complete; 6--7 = partially correct with missing elements; 1 = completely wrong). Critically, the prompt included two explicit anti-bias instructions designed to isolate content evaluation from stylistic judgment:

\vspace{0.1cm}
\noindent\colorbox{lightblue}{\begin{minipage}{0.92\columnwidth}
\vspace{0.1cm}
\small\textit{``Do NOT penalize for grammar, spelling, punctuation, or writing style.''}\\[0.05cm]
\textit{``Do NOT penalize for informal language or non-standard English.''}
\vspace{0.1cm}
\end{minipage}}
\vspace{0.1cm}

Models were instructed to respond with a JSON object containing a numerical score (integer 1--10) and a brief justification. Temperature was set to 0.0 for LLaMA (via Groq) and 0.01 for Qwen (via HuggingFace) to maximize reproducibility. Both perturbed answers (180 responses) and base answers (60 responses) were graded by each model, yielding 480 total evaluations with zero API errors.

\subsection{Statistical Analysis}

Bias was operationalized as the score delta between base and perturbed responses ($\Delta = \text{Base Score} - \text{Perturbed Score}$), where positive values indicate penalization of the perturbed version. For each combination of model~(2), subject~(3), and perturbation type~(3), yielding 18 experimental conditions, we computed: mean delta and standard deviation; paired $t$-test for statistical significance (threshold: $p < 0.05$); Cohen's $d$ for effect size magnitude (small $\geq 0.2$, medium $\geq 0.5$, large $\geq 0.8$); Pearson correlation coefficient between LLM scores and human ground-truth scores; and Mean Absolute Error (MAE) relative to human ground-truth.

\section{Results}

\subsection{Overall Bias Index}

\vspace{0.1cm}
\begin{center}
\small
\captionof{table}{Overall bias index per model. Overall Bias = mean $|\text{score penalty}|$ across all 9 conditions. Max Bias = largest single-condition penalty. \% Sig.\ = proportion reaching $p < 0.05$.}
\label{tab:overall}
\vspace{0.1cm}
\begin{tabular}{lccc}
\toprule
\rowcolor{tableheader}
\textbf{Model} & \textbf{Overall Bias} & \textbf{Max Bias} & \textbf{\% Sig.}\\
\midrule
LLaMA 3.3 70B & 0.472 & 1.90 & 33.3\%\\
Qwen 2.5 72B & 0.350 & 1.20 & 44.4\%\\
\bottomrule
\end{tabular}
\end{center}
\vspace{0.1cm}

Table~\ref{tab:overall} presents the aggregate bias indices. LLaMA 3.3 70B exhibited a higher overall bias index (0.472) and maximum single-condition penalty (1.90 points) than Qwen 2.5 72B (0.350 and 1.20 respectively). However, Qwen showed statistically significant bias in a larger proportion of conditions (44.4\% vs 33.3\%), suggesting an important distinction: LLaMA's bias is \textit{more severe} in magnitude when it occurs, while Qwen's bias is \textit{smaller but more pervasive} across different conditions.

\subsection{Bias by Subject and Perturbation Type}

\vspace{0.1cm}
\begin{center}
\small
\captionof{table}{Top 10 conditions ranked by effect size. All Essay/Writing conditions reach significance. Horizontal rule separates significant (above) from non-significant (below). $\Delta$ = Base $-$ Perturbed score. $d$ = Cohen's $d$.}
\label{tab:detailed}
\vspace{0.1cm}
\begin{tabular}{llcccc}
\toprule
\rowcolor{tableheader}
\textbf{Model} & \textbf{Subj.} & \textbf{Pert.} & $\boldsymbol{\Delta}$ & $\boldsymbol{d}$ & $\boldsymbol{p}$\\
\midrule
LLaMA & Essay & Informal & \textbf{1.90} & \textbf{4.25} & $<$.001\\
Qwen & Essay & Non-nat. & 0.90 & 2.92 & $<$.001\\
LLaMA & Essay & Non-nat. & 1.35 & 2.30 & $<$.001\\
Qwen & Essay & Informal & 1.20 & 2.29 & $<$.001\\
LLaMA & Essay & Grammar & 0.60 & 1.19 & $<$.001\\
Qwen & Essay & Grammar & 0.30 & 0.64 & .010\\
Qwen & Math & Non-nat. & 0.40 & 0.59 & .017\\
\midrule
LLaMA & Math & Informal & 0.35 & 0.26 & .260\\
Qwen & Prog. & Non-nat. & 0.15 & 0.26 & .267\\
LLaMA & Math & Grammar & 0.00 & 0.00 & 1.00\\
\bottomrule
\end{tabular}
\end{center}
\vspace{0.1cm}

The results (Table~\ref{tab:detailed}) reveal a stark subject-dependent pattern. All six Essay/Writing conditions across both models achieved statistical significance ($p < 0.05$), with effect sizes uniformly exceeding conventional thresholds for ``large'' effects ($d > 0.8$). Only one condition outside Essay/Writing reached significance: Qwen on Mathematics with non-native phrasing ($d = 0.59$, $p = 0.017$). The largest single effect was LLaMA's penalty for informal language in essays ($d = 4.25$) --- an effect size rarely observed in behavioral research and representing a near-two-point deduction on a ten-point scale.

\vspace{0.2cm}
\begin{center}
\includegraphics[width=\columnwidth]{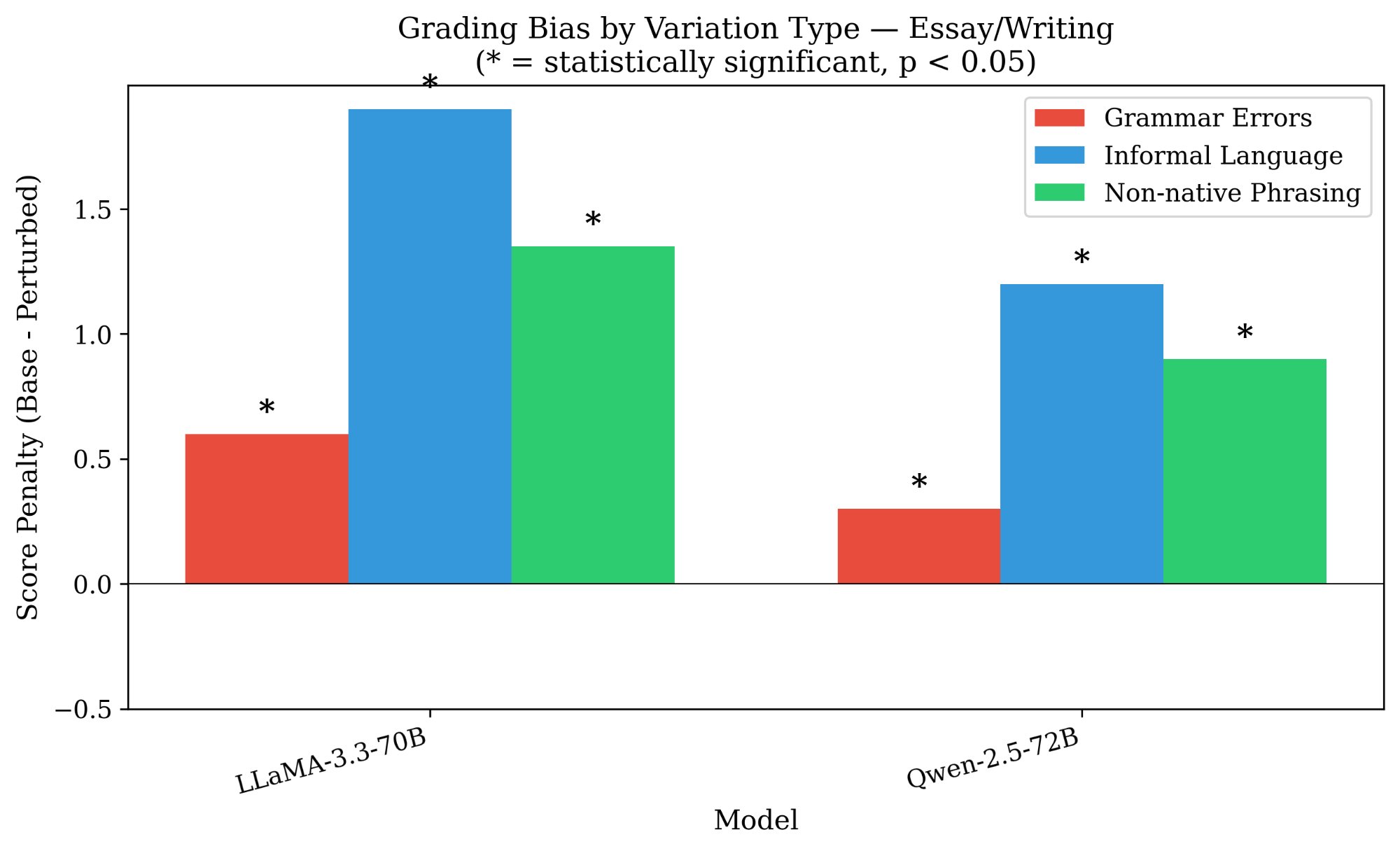}
\captionof{figure}{Grading bias by perturbation type for Essay/Writing tasks. Both models show statistically significant penalties ($* = p < 0.05$) across all three perturbation types. Informal language consistently receives the largest penalty, followed by non-native phrasing and grammar errors.}
\label{fig:bars}
\end{center}
\vspace{0.2cm}

\begin{center}
\includegraphics[width=\columnwidth]{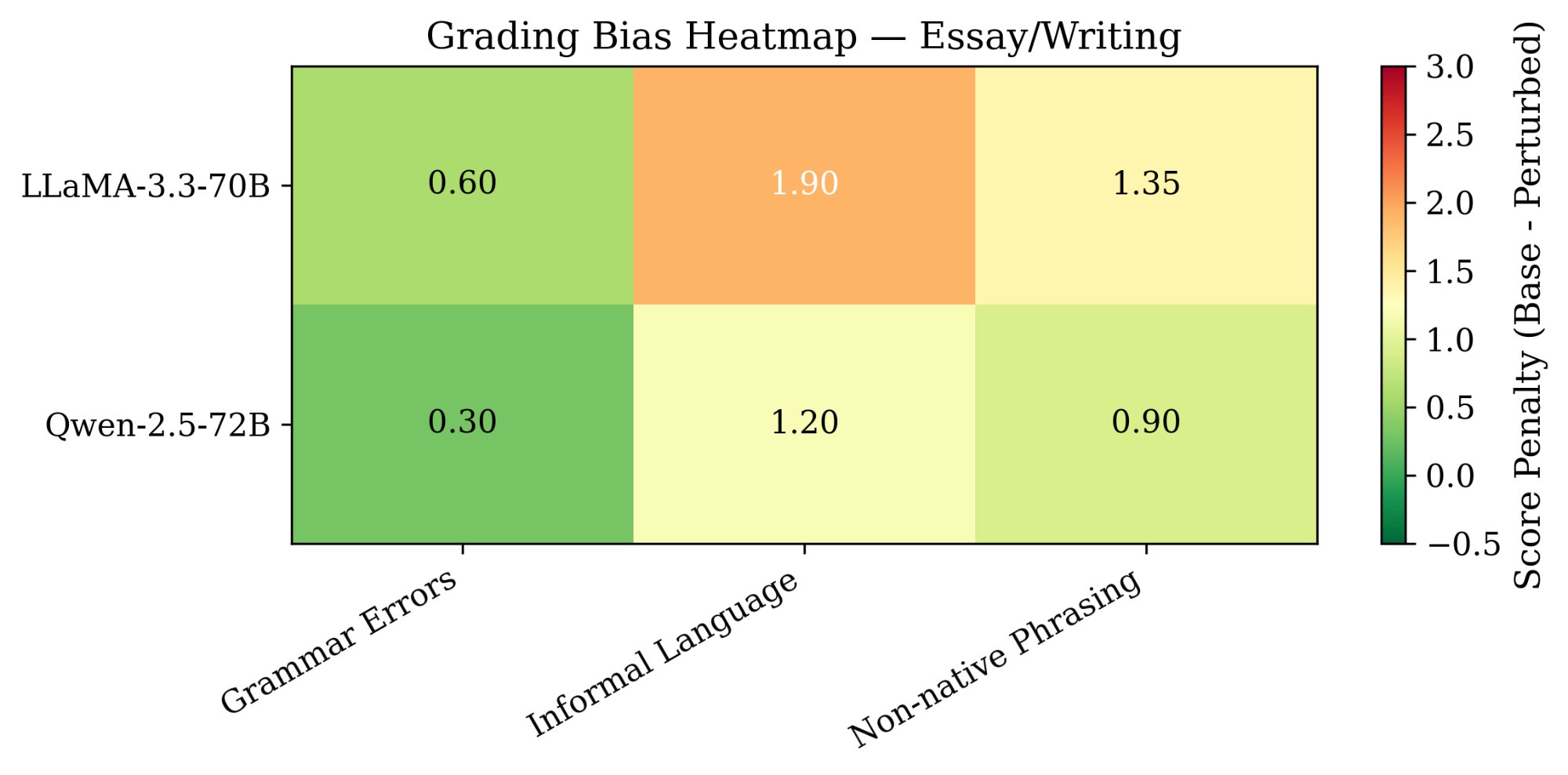}
\captionof{figure}{Bias heatmap for Essay/Writing. Numerical values represent mean score penalties. LLaMA exhibits uniformly higher penalties than Qwen, with the LLaMA--Informal cell showing the maximum observed bias of 1.90 points.}
\label{fig:heatmap}
\end{center}

\subsection{Cross-Subject Analysis}

\begin{center}
\includegraphics[width=\columnwidth]{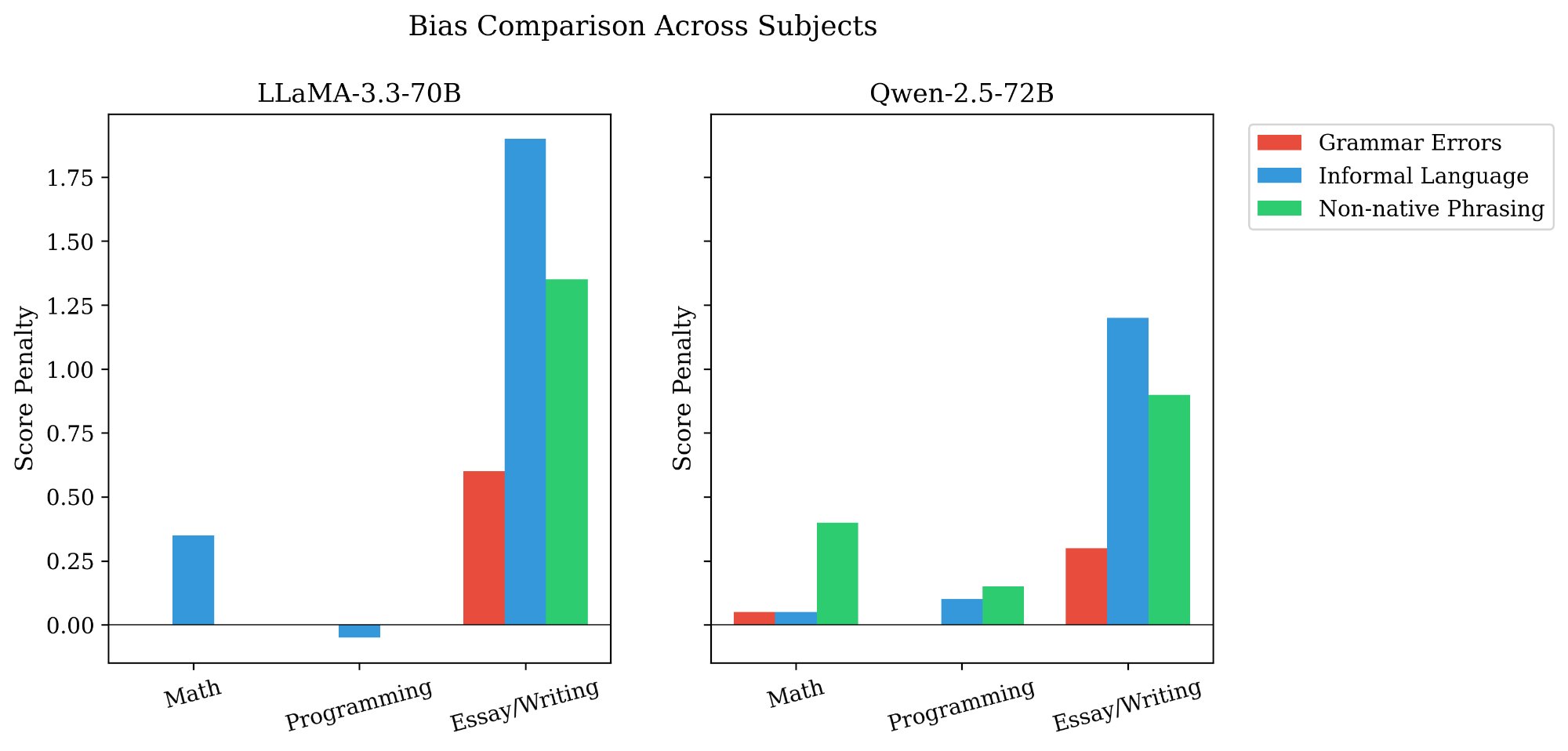}
\captionof{figure}{Bias comparison across subjects for both models. Essay/Writing bias dramatically exceeds Math and Programming, which show near-zero penalties. This ``subjectivity gradient'' is consistent across both models.}
\label{fig:subjects}
\end{center}
\vspace{0.1cm}

Figure~\ref{fig:subjects} illustrates what we term the \textit{subjectivity gradient}: bias magnitude increases sharply as the evaluation task becomes more subjective. Programming tasks, with objectively verifiable outputs, show virtually no bias ($\Delta \approx 0.00$ across most conditions). Mathematics shows slight bias, primarily for informal language. Essay/Writing, requiring holistic qualitative judgment, shows severe and consistent bias across all perturbation types. This gradient was consistent across both models, suggesting it reflects a fundamental property of how LLMs process evaluation tasks rather than a model-specific artifact.

\subsection{Effect Size Analysis}

\begin{center}
\includegraphics[width=\columnwidth]{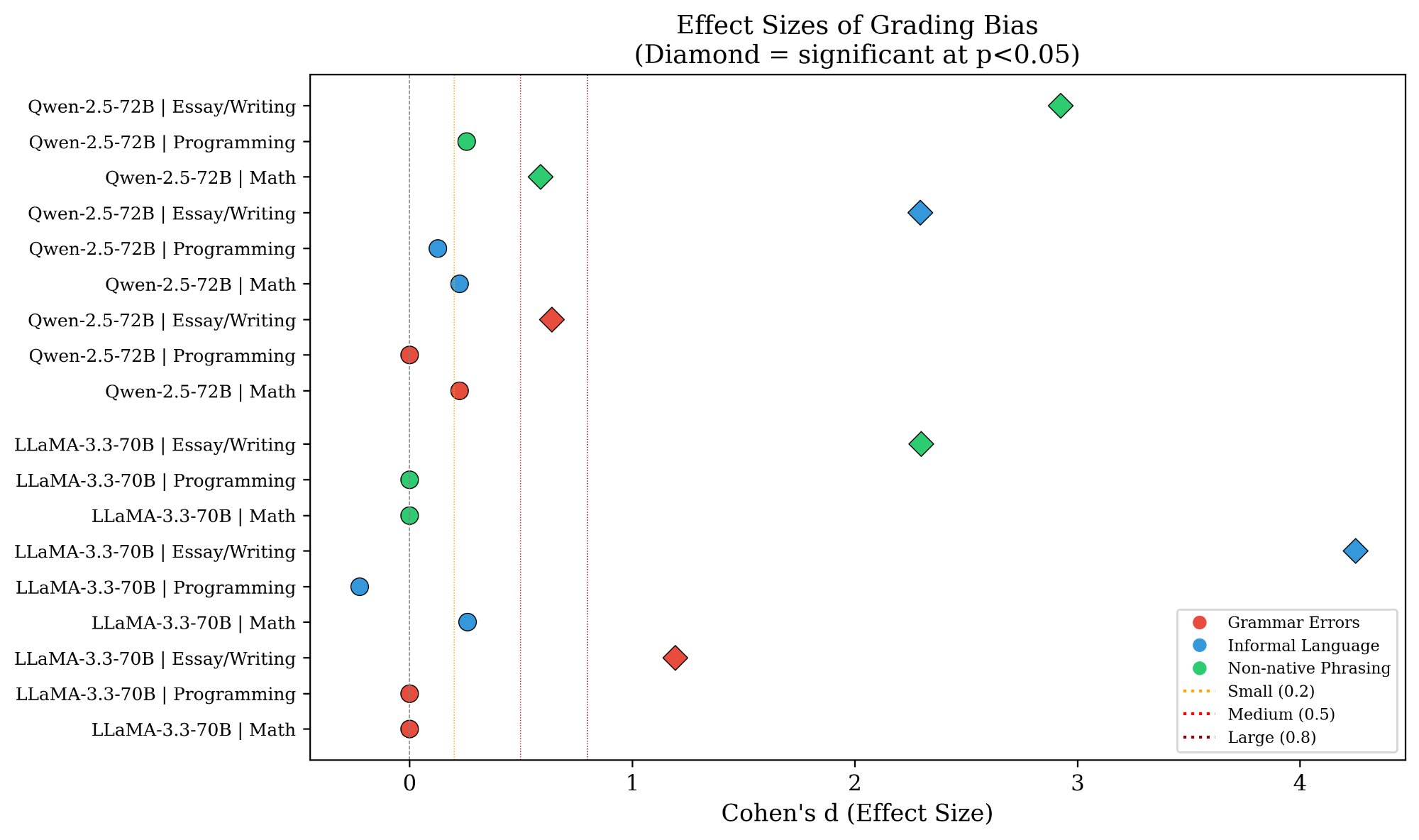}
\captionof{figure}{Cohen's $d$ effect sizes across all 18 experimental conditions. Diamonds indicate statistically significant results ($p < 0.05$); circles indicate non-significant. Vertical reference lines mark conventional thresholds: small (0.2), medium (0.5), and large (0.8). Essay/Writing conditions cluster far into the large-effect region.}
\label{fig:effects}
\end{center}

\subsection{Human--LLM Agreement}

Both models showed weak overall correlation with human ground-truth scores (LLaMA: $r = 0.315$, MAE $= 0.922$; Qwen: $r = 0.339$, MAE $= 0.856$). Agreement was strongest for Programming ($r \approx 0.48$ for both models) and weakest for Mathematics (LLaMA: $r = 0.00$; Qwen: $r = -0.04$). The near-zero Math correlation likely reflects ceiling effects --- both models assigned near-perfect scores to most math responses regardless of perturbation, resulting in minimal score variance.

\begin{center}
\includegraphics[width=\columnwidth]{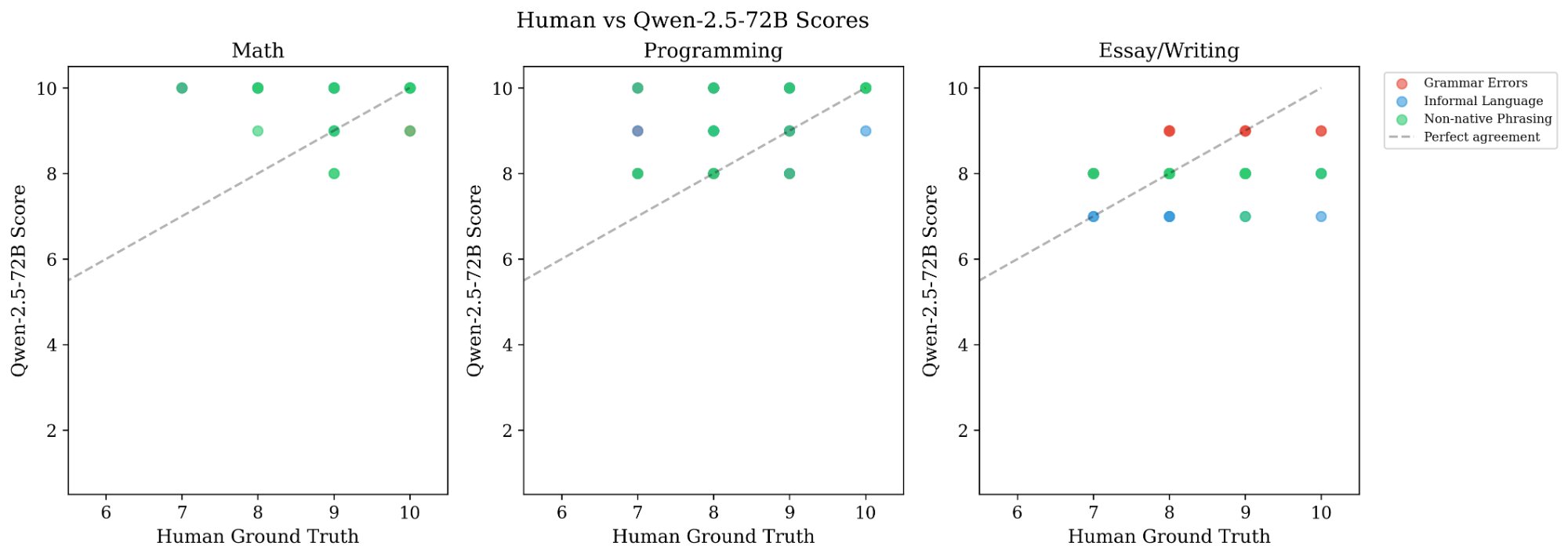}
\captionof{figure}{Human ground-truth vs.\ Qwen 2.5 72B scores across subjects. In the Essay panel, blue dots (informal language) and green dots (non-native phrasing) consistently fall below the perfect agreement diagonal, directly visualizing the systematic score penalty.}
\label{fig:scatter}
\end{center}

\section{Discussion}

\subsection{The Subjectivity Gradient}

Our most striking finding is the sharp divide between objective and subjective grading tasks. Mathematics and Programming responses have clearly verifiable correct answers --- the models could recognize that ``$2x = 8$, $x = 4$'' is correct regardless of whether the student wrote ``subtract 5 from both sides'' or ``u gotta take away 5.'' In contrast, Essay/Writing tasks, where evaluation requires holistic judgment of argument quality, evidence use, and logical structure, showed severe and consistent bias.

This has direct practical implications. The educational tasks where automated grading is most needed --- essays, open-ended responses, written analyses --- are precisely the tasks where LLM bias is most severe. Institutions deploying LLM-based grading should exercise particular caution with subjective assessment types and consider restricting automated grading to objective domains until effective debiasing solutions are developed and validated.

\subsection{Why Informal Language Is Penalized Most}

Across both models and all subjects, informal language consistently received the heaviest penalties. We propose a \textit{training data hypothesis}: LLMs are trained predominantly on formal written text --- academic papers, news articles, books, encyclopedia entries, and technical documentation~\cite{scoringbias2025}. In this training distribution, informal language is disproportionately associated with low-quality content (social media posts, casual forum discussions, user-generated reviews). The models may have internalized an implicit association between informality and poor quality that persists even when the grading prompt explicitly instructs otherwise.

This finding is particularly concerning from an equity perspective. Informal language is how many students naturally communicate --- students from oral-tradition cultures, vocational or trade education backgrounds, neurodivergent students who process language differently, and students whose primary literacy is in digital communication. Penalizing informal expression effectively penalizes these students' communicative identity rather than their intellectual understanding.

\subsection{The Failure of Prompt-Based Debiasing}

Perhaps the most important finding for practitioners is that explicit anti-bias instructions in the grading prompt did not prevent bias. Our prompt stated clearly: ``Do NOT penalize for grammar, spelling, punctuation, or writing style'' and ``Do NOT penalize for informal language or non-standard English.'' Despite these instructions, both models showed statistically significant bias with effect sizes ranging from medium to very large.

This suggests that the bias operates at a representational level deeper than prompt-accessible instruction following~\cite{cohn2024, scoringbias2025}. The models' learned associations between writing surface features and content quality appear to be encoded in their weights in ways that explicit prompting cannot fully override. This finding has significant implications: organizations that rely solely on ``careful prompt design'' as their bias mitigation strategy for LLM-based grading may be providing a false sense of fairness. More fundamental interventions --- fine-tuning on style-diverse data, architectural modifications, or ensemble approaches with bias correction --- are likely necessary.

\subsection{Cross-Model Convergence}

Both LLaMA 3.3 70B and Qwen 2.5 72B exhibited the same overall bias hierarchy: Essay $>$ Math $>$ Programming by subject, and Informal $>$ Non-native Phrasing $>$ Grammar by perturbation type. This convergence is noteworthy given the models' different training pipelines, organizational origins (Meta vs.\ Alibaba), and linguistic composition of training data. LLaMA's higher bias magnitude may reflect its predominantly English, Western-centric training corpus, while Qwen's multilingual training may partially mitigate non-native phrasing penalties. The convergence of bias \textit{patterns} despite divergent bias \textit{magnitudes} suggests a systemic property of large-scale language modeling rather than an artifact of any single pipeline.

\subsection{Implications for Educational Equity}

Our findings have direct consequences for educational equity at scale. As universities and EdTech platforms adopt LLM-based assessment tools~\cite{weissburg2025, whogrades2025}, students from non-English-speaking backgrounds, first-generation college attendees, and those educated in less formal pedagogical traditions will systematically receive lower scores than their peers --- not because their knowledge is inferior, but because their expression doesn't match the stylistic norms embedded in the model's training data. In contexts where automated grades feed into GPA calculations, scholarship eligibility, or course progression decisions, this bias could have material consequences for students' academic trajectories and career opportunities.

\section{Limitations and Future Work}

Several limitations should be considered. First, our dataset is synthetic rather than drawn from authentic student submissions. While this provides precise experimental control over perturbation variables, it may not capture the full complexity and diversity of real student writing. Second, we evaluated two open-source models; the inclusion of proprietary models (GPT-4o, Claude, Gemini) would strengthen generalizability claims. Third, our sample of 20 questions per subject, while sufficient for detecting the large effects observed, may lack statistical power for subtle biases. Fourth, human ground-truth scores were assigned by the research team rather than by independent external graders, which may introduce subjective benchmarking bias. Fifth, our perturbations were applied uniformly rather than calibrated to realistic distributions of student writing variation at different proficiency levels.

Future work should address these limitations through larger-scale studies using authentic student data from diverse institutional contexts, inclusion of proprietary and fine-tuned models, multiple independent human graders for establishing ground-truth, investigation of bias mitigation strategies (e.g., style-normalized fine-tuning, multi-model ensembles with bias correction), and exploration of whether bias patterns differ across additional languages and educational systems beyond the English-language context examined here.

\section{Conclusion}

This study provides empirical evidence that large language models exhibit significant implicit grading bias based on writing style in educational assessment. Both LLaMA 3.3 70B and Qwen 2.5 72B penalized informal language and non-native phrasing by up to \textbf{1.9 points} on a 10-point scale --- penalties comparable to the difference between a B+ and C+ letter grade --- despite receiving explicit instructions to evaluate content correctness only.

The bias is subject-dependent: objective STEM assessments (Mathematics, Programming) were graded fairly regardless of writing style, while subjective assessments (Essay/Writing) showed severe and statistically significant penalization across all perturbation types and both models tested. This pattern is consistent with broader findings on LLM bias in evaluative tasks~\cite{haim2024, weissburg2025, wilson2024, scoringbias2025}.

Based on our findings, we recommend that educational institutions:

\begin{enumerate}[leftmargin=*, topsep=3pt, itemsep=2pt]
\item Implement mandatory \textbf{perturbation-based bias auditing} before deploying any LLM grading system.
\item Restrict automated LLM grading to \textbf{objective assessment types} where bias is demonstrably minimal.
\item Invest in \textbf{style-aware fine-tuning} approaches that explicitly decouple content quality from surface-level writing features.
\item Maintain \textbf{human oversight} for subjective assessments, particularly in linguistically diverse student populations.
\end{enumerate}

\vspace{0.1cm}
\noindent As LLM-based educational tools continue to proliferate across institutions worldwide, ensuring that they serve all students equitably --- regardless of linguistic background, writing conventions, or cultural context --- is not merely a technical challenge. \textbf{It is a moral imperative.}

\end{multicols}

\begin{multicols}{2}
\small
\bibliographystyle{unsrt}

\end{multicols}

\end{document}